# *HCDIR: End-to-end <u>H</u>ate <u>C</u>ontext <u>D</u>etection, and <u>I</u>ntensity <u>R</u>eduction model for online comments*

Neeraj Kumar Singh[1†], Koyel Ghosh[2*†], Joy Mahapatra[1], Utpal Garain[1], Apurbalal Senapati[2]

[1]CVPRU, Indian Statistical Institute, Kolkata, 700108, West Bengal, India.
[2]CSE, Central Institute of Technology, Kokrajhar, 783370, Assam, India.

*Corresponding author(s). E-mail(s): ghosh.koyel8@gmail.com;
Contributing authors: neeraj1909@gmail.com;
joymahapatra90@gmail.com ; utpal.garain@gmail.com;
a.senapati@cit.ac.in;
†These authors contributed equally to this work.

**Abstract**

**Warning: This paper contains examples of the language that some people may find offensive.**
Detecting and reducing hateful, abusive, offensive comments is a critical and challenging task on social media. Moreover, few studies aim to mitigate the intensity of hate speech. While studies have shown that context-level semantics are crucial for detecting hateful comments, most of this research focuses on English due to the ample datasets available. In contrast, low-resource languages, like Indian languages, remain under-researched because of limited datasets. Contrary to hate speech detection, hate intensity reduction remains unexplored in high-resource and low-resource languages. In this paper, we propose a novel end-to-end model, **HCDIR**, for **H**ate **C**ontext **D**etection, and **H**ate **I**ntensity **R**eduction in social media posts. First, we fine-tuned several pre-trained language models to detect hateful comments to ascertain the best-performing hateful comments detection model. Then, we identified the contextual hateful words. Identification of such hateful words is justified through the state-of-the-art explainable learning model, i.e., Integrated Gradient (IG). Lastly, the Masked Language Modeling (MLM) model has been employed to capture domain-specific nuances to reduce hate intensity. We masked the 50% hateful words of the comments identified as hateful and predicted the alternative words for these masked terms to generate convincing sentences. An optimal replacement for the original hate comments from the feasible sentences is preferred. Extensive experiments have been conducted on several recent datasets using automatic metric-based evaluation (BERTScore) and thorough human evaluation. To enhance the faithfulness in human evaluation, we arranged a group of three human annotators with varied expertise.

# 1 Introduction

The current surge in social media usage has resulted in the widespread availability of hateful content. Cambridge Dictionary[1] defines the word 'Hate speech' as—"a public speech that expresses hatred or encourages violence against a person or group based on race, religion, gender, or sexual orientation". Recent studies reveal that half of the world's population, including print media, are now engaged in social media platforms [1], and 12.5 trillion hours are spent online by users [2]. This trend will undoubtedly continue indefinitely, creating opportunities for user feedback, but aggressive posts, false news, and harassing comments can all lead to social violence, riots, and other forms of retaliation [3]. Worldwide, governments are introducing laws against hate speech. So, digital media like ™Twitter, ™Facebook, etc., are also becoming more concerned about it and endeavoring to filter hate, sexual abuse, harmful acts, harassment, bullying, child abuse, etc. United Nations Strategy and Plan of Action on Hate Speech [4] has been introduced.

Researchers explore the field to find a solution to detecting hate speech. Most of the work in hate speech detection has been done in high-resource languages such as English[2] [5–8]. In the case of low-resource Indian languages, much work has been done on hate speech detection. The introduction of large language models (LLMs) led to a significant improvement in our proposed method's performance for downstream tasks like classification, summarization, and visual question answering. However, these models still lag in giving proper insights about their decision-making. Therefore, it becomes hard for the end-users to understand the proper reasoning behind the model decision. Few attempts have been made to explain the hate speech detection model's behaviour. Mathew et al. [9] observed that models that perform very well in classification do not score highly on explainability metrics like plausibility and faithfulness. Rezaul et al. [10] proposed an explainable approach for hate speech detection in Bengali. They have tried Sensitive Analysis (SA) and Layerwise Relevance Propagation (LRP) on a deep network baseline model and listed the top features in each category-religious, geo-political, political, and personal. Apart from hate detection and hate explanation, little research has been done on hate intensity reduction. Tanmoy et al. have [11] tried 'hate speech normalization' to weaken the intensity of hatred in an online post. In their work of parallel corpus, hate spans, and their normalized counterparts, they reduced hate by paraphrasing the hate spans.

---

[1]https://dictionary.cambridge.org/dictionary/english/hate-speech
[2]https://hatespeechdata.com/

In this paper, we have fine-tuned the three different transformer-based models, Google-MuRIL [12] based on BERT, XLM-R [13] based on RoBERTa, and Indic-BERT [14] based on ALBERT for detecting the hate content in the social media post and produce comparisons of the results generated by these three models. Then, the system finds the words of the input instance that are influencing the model prediction using the explainability approach. Ultimately, Masked Language Modelling(MLM) is used to reduce the hate content in hate speech. The detected hate words have masked and replaced with the most suitable ones using metrics like BERTScore [15].

Our main focuses are:

1. Detection of hate context of a given comment
2. Explain the model prediction using available explanation methods.
3. Reducing the hate intensity in the hate post

The rest of the paper is structured as follows. Section 2 is the work on detecting hate context and reducing hate intensity in Indian languages. Section 3 explains the objectives, and Section 4 discusses the proposed method. Section 5 is dedicated to the Experiments. Section 6 discusses error analysis. Section 7 illustrates the conclusion.

## 2 Related Work

In this section, we discuss related works in two parts- 1) Hate context detection and 2) Hate intensity reduction.

### 2.1 Hate context detection

BOW is another model similar to dictionaries is the bag-of-words [16–18]. Instead of using a predetermined collection of terms like dictionaries, a corpus is constructed using the words found in the training data. After collecting all words, their frequency is utilized to train a classifier. The disadvantage is this method ignores word sequence and syntactic and semantic meaning. One of the most popular methods for automatic hate speech detection and related tasks is using N-grams [19–24]. Assembling lists of N words by connecting sequences of words is the most typical N-gram method. In this scenario, we tally up all the occurrences of expressions of size N. As a result, the performance of classifiers can be enhanced by using this information to understand the meaning of individual words better. N-grams can also be used with characters or syllables in place of words. This method is less affected by occasional changes in spelling than others. Regarding abusive language identification, character N-gram features were more predictive than word N-gram features [25]. There are, however, drawbacks to using N-grams. A potential drawback is that even closely related words can be far apart in a sentence [20], and solutions to this issue, like increasing the N value, slow down processing performance [26]. Research also shows that a larger N ($\geq 5$) improves performance over smaller N values (1 or 3 gram) [23]. According to a survey [27], N-gram features are widely regarded as highly predictive in the problem of automatic detection of hate speech. However, they show the most promise when used with other characteristics. TF-IDF (term frequency-inverse document frequency) is also applied to sentiment classification problems [28]. The TF-IDF determines how significant a word is inside a corpus of documents. To account for the fact that some words appear more frequently than others in the corpus, the frequency of the term is off-setted by the frequency of the word in the corpus, setting it apart from an N-gram bag of words (e.g., stop words). Using a paragraph2vec technique, Djuric et al. [29] can determine whether a user's comment contains abusive or appropriate language and guess the message's main word using FastText [19] embeddings. Sentences, not individual words, must be categorized, which presents a challenge for hate speech identification [27]. The problem can be solved by averaging the word vectors in the sentence. This approach, however, only works up to a certain point [30]. Some authors suggest comment embeddings get around this issue [31]. A recurrent neural network (RNN) is a neural network that uses the results of the previous phase as input for the one being performed now. Saksesi et al. [32] use RNN to classify hate speech in the text. The Disadvantage of RNN is that the gradient descent problem does not improve accuracy after a certain time for the longer texts. Das et al. used long short-term memory networks (LSTM) [33]. Anand et al. [34] use LSTM and CNN with and without word embedding to classify abusive comments. Bashar et al. [35] used Word2Vec word embedding with Convolutional Neural Network (CNN) to classify hate comments in the Hindi language [36] and scored the highest macro F1 score. Raj et al. [37] scored the highest macro F1 score utilizing CNN and Bidirectional Long Short-Term Memory (BiLSTM) in a hate speech shared task [38]. In another hate speech Hindi shared task [39], the best submission was achieved the highest Macro F1 fine-tuning Multilingual-BERT. In the paper [40], they used the pre-trained multilingual BERT (m-BERT) model for computing the input embedding on the Hostility Detection Dataset (Hindi) later SVM, Random-Forest, Multilayer perceptron (MLP), Logistic Regression models are used as classifiers. In coarse-grained evaluation, SVM reported the best-weighted F1 score.

## 2.2 Explainable Method

Attempts to define the idea of explainability have met with limited success. Early works in model interpretation started with Inherently Interpretable models, like rule-based methods, risk scores, prototype-based models, attention-based models, etc. Rule-based models use the sequences of if/else-if/else rules. A recent work is Rule-based lists [41], which results in a probabilistic classifier that uses a mixture of theoretical bounds to optimize over rule lists fully. Risk scores are widely used in medicine and criminal justice, allowing the decision-makers to understand the results easily. Recent works [42] use the Risk scores to detect the 'at-risk' patents in a real-time environment. Anphi et al. [43] introduced monotonicity between features and outputs to reason about a single feature on an output independently from other features. In recent work in prototype-based models [44], authors created a novel architecture by adding a special prototype layer in the deep network that naturally explains its reasoning for each prediction. The prototypes are learned during training;

the learned network naturally explains each prediction. Another work by [45] uses hierarchically organized prototypes to classify the objects at every level in a predefined taxonomy, giving distinct explanations at each level in a predefined taxonomy. Attention-based models are first introduced by Bahdanau et al. [46]. They extended the idea of a fixed-length vector in neural machine translation to automatically search for the parts of a source sentence relevant to predicting a target word. Ashish et al. [47] introduced the idea of a self-attention mechanism relating different positions of a single sequence to compute a representation of the sequence. Self-attention has been used successfully in various tasks, including reading comprehension, abstractive summarization, textual entailment, and learning task-independent sentence representations.

Several post-hoc-based explanation algorithms [48] have been introduced to explain the black-box nature of deep-learning models. One such technique is LIME [49], which can explain any classifier's predictions in an interpretable and faithful manner. Another approach is SHAP [50], which calculates the Shapley value using multiple observations. The authors have presented a unified framework for interpreting the predictions. There are several feature feature attribution algorithms that use saliency maps, such as Layer-Wise Relevance Propagation (LRP) [51], DeepLIFT [52], Guided Backpropagation [53], and many others. LRP allows the decomposition of the prediction of a deep neural network computer over a sample, like a text, into a relevance score for input embeddings. DeepLIFT(Deep Learning Important Features) breaks down the output prediction of a network on a specific input by backpropagating the contributions of all the neurons in the network to every feature of the input. It compares the activation of each neuron to its reference activation and assigns contribution scores according to the difference. Guided Backpropagation combines vanilla backpropagation at ReLUs [54] and DeconvNets [55]. It visualizes the gradient for the input image to capture pixels detected by the neurons.

Sundararajan et al. [56] studied the problem of attributing the prediction of a deep network to the input features. They identified two axioms- sensitivity and implementation invariance, that attribution methods should satisfy. They showed that the most well-known attribution methods do not satisfy these axioms. They used this axiom to design the new attribution method called IG. The authors [57] present a novel exemplar-based contrastive learning approach, i.e. the Rule By Example (RBE) for learning from logical rules for textual content moderation.

## 2.3 Hate intensity Reduction

To reduce the severity of hate speech, a novel task, the hate intensity reduction task, is introduced to reduce the hatred of an online post. Sarah Masud et al. [11] propose hate speech normalization to provide users with a stepping stone towards non-hate.They have manually created a parallel corpus of the hate texts and their counterparts (less hateful). They introduced a hate speech normalization model that has three stages— stage 1 measures the hate intensity of the original post; stage 2 identifies the hate span(s) within it; and finally, stage 3 reduces the hate intensity by paraphrasing the hate spans.

## 3 Objective

There are two key objectives of our proposed method—(1) Hate Context Detection (HCD); and (2) Hate Intensity Reduction (HIR). Consider an input instance $x$ consisting of $n$ tokens. We aim to detect whether the input is belongs to hate text $h$ and obtain their normalized form $h'$. So, to achieve this, the overall task is divided into two stages:

- **Hate Context Detection (HCD)**: In this phase, we detect hate texts $h$ among $x$. Then, we identify the top contextual phrases/ words $h_w$, which influence words responsible for the model prediction. i.e., $h_w$ = HCD(x).
- **Hate Intensity Reduction (HIR)**: Now, we have top contextual phrase/ words $h_w$, and the hate text $h$, then reduce hatred from the text. i.e., $h'$ = HIR($h_w$, $h$)

## 4 Proposed Method

We intend to design a complete end-to-end system that detects hate comments as well as identifies the most hateful ones and reduces the amount of hatred in comments. The architecture of the system is straightforward and beneficial compared to any complex system. Figure-1 illustrates a high-level overview of our **HCDIR** (**H**ate **C**ontext **D**etection and **I**ntensity **R**eduction) framework. HDIR framework will be called into action once the system receives an input instance $x$. The first step is to detect any hate comments from the input. Once a hate comment is identified, the words with the positive attribution scores from the comment are identified. Finally, the hate in the comments is reduced, and it is transformed into a new, less hateful comment $h'$.

### 4.1 Hate context detection

The objective of this task is to determine whether a given comment is *HOF (Hate or Offensive speech)* or *NOT*. Our dataset, denoted as D, comprises p comments, represented as $\{x_1, x_2, x_3, ..., x_i, ..., x_p\}$, where $x_i$ represents the i-th comment, and p is the total number of comments in the dataset. Each comment $x_i$ has m words, which are denoted as $x_i = \{w_1, w_2, w_3, ..., w_k, ..., w_m\}$, where $w_k$ represents the k-th word din the comment. The dataset can be defined as $D = \{(x_1, y_1), (x_2, y_2), (x_3, y_3), ..., (x_i, y_i), ...., (x_p, y_p)\}$, where each tuple consists of the comment ($x_i$) and its corresponding label ($y_i$). The label indicates whether the comment is HOF or NOT. In other words, $D = \{(x_i, y_i)\}_{i=1}^{p}$, where $y_i$ is either HOF or NOT. This task is a binary classification problem, and the goal is to maximize the objective function

$$\arg\max_{\theta}(\Pi_{i=1}^{p}(P(y_i|x_i;\theta))) \quad (1)$$

where $x_i$ is a comment with a tagged label $y_i$ to be predicted. $\theta$ are the model parameters that require optimization. The process involves creating a classifier to categorize comments into two classes. To achieve this, the training dataset is divided into training and validation sets. The aim is to train the classifier on the training set and then evaluate its performance on the validation set. The focus is on converting

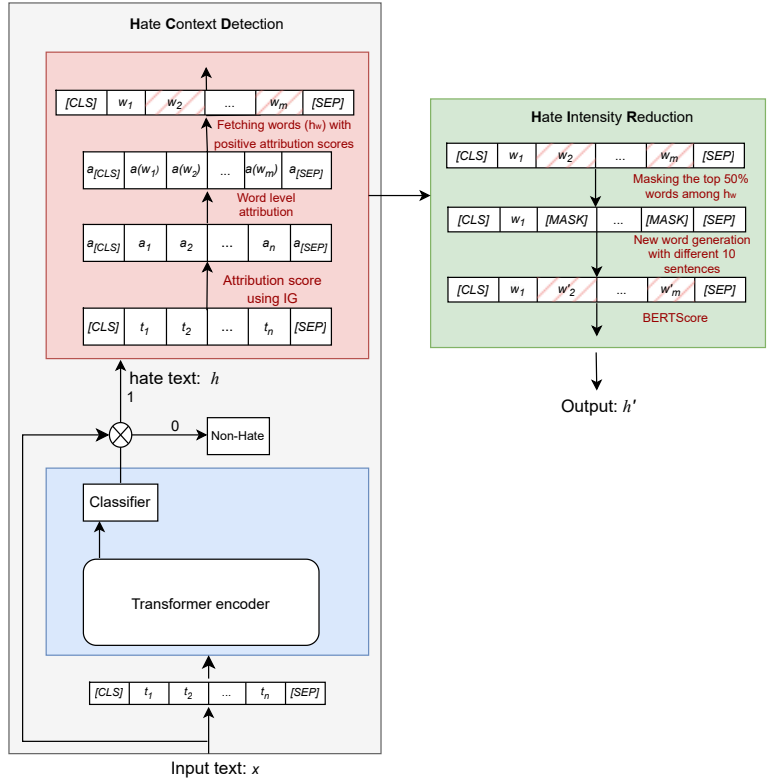

**Figure 1** This is the end-to-end architecture of HCDIR. Here, the HCD part detects hate text and contextually influenced hate words; later, hate words and hate text are used as input to the part of the HIR to produce the resultant words (box crossed with slanted lines)

the comment or text data into a format that a machine can understand. The model distinguishes between the two classes by analyzing the processed text data. During the learning phase, the algorithm adjusts its internal settings based on the training data, improving its ability to make accurate predictions. These adjustments are guided by the differences between the two classes in the training data. The outcome of this learning process is a classifier that can take new text data as input and predict which of the two classes it belongs to. This classifier is then tested using the test set to see how well it performs on unseen examples. The process involves training a system to understand and categorize text by learning from examples, which is then tested on new unseen text to ensure reliability. The tokenization form of the input instance $x$ to a transformer-based model is $x = \{[CLS], t_1, t_2, t_3, ...., t_n, [SEP]\}$, where $t_n$ is the $n^{th}$ token in $x$. We calculate the contextual embeddings $e$ using pre-trained transformer models.

We extract the embedding for the ['CLS'] token from the contextual embedding ($e$). This ['CLS'] embedding is passed through a dense layer with a softmax activation to determine the probability of each sentiment class.

$$e = \text{BERT\_Encoder}(x) \qquad (2)$$
$$p = \text{softmax}(Linear(e['CLS'])) \qquad (3)$$
$$c = \text{argmax}(p) \qquad (4)$$

In the equations above, BERT_Encoder in equation 1 represents the BERT encoding process, and Linear in equation 2 denotes the classification layer that yields the probability scores for each class.

In equation 3, if the value of c is equal to 1, our next target is to detect the context that influences a model to classify a comment as a hate comment. To achieve this, we employ post-hoc analysis in our experimental framework. We use IG to calculate the attribution score. For the input instance represented by $t$ and the baseline input represented by $\tilde{t}$, we assess the contribution of each token embedding to the model's predicted class probability. This allows us to determine the attribution for each component of the embedding vector and compute the overall attribution of a token by averaging these values. The formula used to calculate the IGs is given by,

$$IG(t, \tilde{t}) = (t - \tilde{t}) \times \sum_{\alpha=0}^{1}[\nabla F(\tilde{t} + \alpha \times (t - \tilde{t}))]$$

where $t$ is input text, $\tilde{t}$ is baseline text, and $\alpha$ is a scalar in the range [0, 1].

Using the HCD module, we detect whether a text is hateful. If it is hateful, we calculate the attribution score $IG(t, \tilde{t})$ for each token i.e. $\{a_{[CLS]}, a_1, a_2, ...., a_n, a_{[SEP]}\}$. As the sub-words have no semantic sense, we are summing the attribution score of the sub-words to calculate the attribution score for each word, i.e., $\{a_{[CLS]}, a(w)_1, a(w)_2, ...., a(w)_n, a(w)_{[SEP]}\}$. Finally, we get the top-k attribution-scored words $h_k$. Later, $h_k$ and the $h$ are taken as input to the HIR module.

### 4.1.1 Preprocessing

Before passing the data to the transformer model for finetuning, we first preprocess it to improve its quality and performance by removing any noise. The transformer model does not require extensive preprocessing, so we perform basic preprocessing on the text. This includes removing any hyperlinks or links that start with `www`, `http`, or `https`, mentions that start with '@', newline characters, and most punctuation marks (except full stops, commas, and question marks). If necessary, we also convert class labels to unique numerical values.

### 4.1.2 Transformer Language Model

We utilize WordPiece [58] tokenization and dynamic embeddings for improved contextualization. Our study employs three types of pre-trained transformer models:

- **Multilingual model:** Google-MuRIL [12] is a BERT model pre-trained on 17 Indian languages and their transliterated counterparts, including monolingual and parallel segments.

- **Monolingual model:** l3cube-pune/hindi-bert-v2 [59], l3cube-pune/marathi-roberta [60], l3cube-pune/assamese-bert [59], l3cube-pune/tamil-bert [59], l3cube-pune/telugu-bert [59] and l3cube-pune/bengali-bert [59] are used to see monolingual performance o the datasets.

The primary question is, how does the model arrive at a specific conclusion (whether a given input text is hateful or not)? As the model becomes more complex, it becomes increasingly difficult for users to comprehend how it works. For this purpose, we employed the integrated gradients in three stages:

- Assigning an attribution score using IG
- Assigning word-level attribution from subwords
- Choosing the words with positive attribution score.

The transformer-based BERT model has 110 million parameters. To make these Large Language Models (LLMs) more transparent and trustworthy for end users, it's crucial to provide explanations alongside their predictions. Models can be explained in two primary ways: through an inherently explainable approach and a post-hoc explainable approach. In an inherently explainable approach, machine learning models are designed to be transparent and interpretable from the outset. These models, such as decision trees, rule-based models, and linear models, are easier to understand and interpret. On the other hand, post-hoc explanation techniques like IG explain the machine learning model's decisions or predictions.

### 4.1.3 Integrated Gradients (IG)

IGs leverage two key inputs: (a) the input text and (b) the baseline text. The baseline text is carefully chosen to ensure that the model's output is neutral at that point. It serves as a reference input, a starting point to calculate the attribution of each feature to the model's prediction. In this case, we used a sequence of padding tokens as the baseline. We then generated a series of text inputs that gradually move from the baseline to the input text. For each intermediate text input, we calculated the gradient of each layer with respect to the preceding one, moving from the classifier layer to the input layer in sequence. We used the chain rule to determine the gradient of the model's predicted class probability for the input text. We then summed these gradients across the entire path from the baseline to the input text, yielding an attribution value for each entry in the embedding vector. To determine the attribution for the tokens of the input text, we averaged the attributions related to their embeddings. A positive attribution value for a feature indicates that it contributed to the model's predicted class, while a negative value suggests it acted against it. A zero attribution value implies the feature didn't significantly influence the model's prediction for that particular input. Removing or altering such a feature would likely not significantly affect the model's output.

### 4.2 Hate Intensity Reduction

Our goal is not to condone hate context but to provide a way for users to move away from it. To achieve this, we use the HCD module to identify hate texts in all input $x$, spot contextually influenced hate words $h_w$ and replace the hateful words with a ['MASK'] token, creating a masked version of the text, $h'$. We then use Masked Language Modeling (MLM) to predict potential alternatives for the masked words. For each masked text, we generate 10 potential alternative texts. We evaluate these texts using metrics such as BERTScore to select the most promising substitute, denoted as $h'$. This approach helps us reduce the intensity of hate speech within the content.

BERTScore is a metric that evaluates text quality by using contextual embeddings from BERT in the following way:

1. Extracts embeddings for each token in a sentence.
2. Computes cosine similarity between tokens of reference and candidate sentences.
3. Calculates token-level precision, recall, and F1-score.
4. In removing hate speech, the original hateful sentence is the reference, and the substituted sentences are candidates. BERTScore ensures candidates maintain semantic similarity to the reference.
5. It's more nuanced than simple token overlap metrics like BLEU.
6. However, while it measures contextual similarity, it doesn't explicitly gauge the "hatefulness" of a sentence.

## 5 Experiments

### 5.1 Datasets

In this section, we discuss publicly available datasets in Hindi, Marathi, Telegu, Tamil, Bengali, Assamese, and Bodo used for hate speech detection and intensity reduction, respectively. For our experiment, we are using the HASOC (Hate Speech and Offensive Content Identification) dataset [36, 38, 39], and BD-SHS [61]. Statistics and class distributions for the dataset of all HASOC, MACD data, and BD-SHS are given in table 1.

| Dataset | Hate | | | Non-hate | | | Total |
| --- | --- | --- | --- | --- | --- | --- | --- |
| | **Train** | **Val** | **Test** | **Train** | **Val** | **Test** | |
| $HASOC(2019)_{Hindi}$ | 2,469 | - | 605 | 2,196 | - | 713 | 5,983 |
| $HASOC(2021)_{Hindi}$ | 669 | - | 207 | 1,205 | - | 418 | 2,499 |
| $MACD_{Hindi}$ | 10,527 | 3,464 | 3,496 | 9,656 | 3,264 | 3,232 | 33,639 |
| $MACD_{Tamil}$ | 8,302 | 2,824 | 2,795 | 9,698 | 3,176 | 3,205 | 30,000 |
| $MACD_{Telegu}$ | 9,310 | 3,099 | 3,194 | 8,690 | 2,901 | 2,806 | 30,000 |
| $BD-SHS_{Bengali}$ | 19,324 | 2,416 | 2,416 | 20,900 | 2,612 | 2,613 | 50,281 |

**Table 1** Class-wise distribution analysis for HASOC, MACD, and BD-SHS datasets

**Algorithm 1** HCDIR: End-to-end Hate Context Detection, and Intensity Reduction model

---

**Input:** A set of samples $D = \{(x_1, y_1), (x_2, y_2), (x_3, y_3), ..., (x_i, y_i), ...., (t_p, y_p)\}$, where each tuple consists of the text ($t_i$) and label ($y_i$) corresponding to the text.

**Output:** Hate sentences $H = \{h_1, h_2, h_3, ..., h_i, ...., h_p\}$ with the IG-scored words $A = \{a[CLS], a(w)_1, a(w)_2, ...., a(w)_n, a[SEP]\}$.

  **procedure** HCDIR($D$)
    $A \leftarrow$ HCD($D$)                                                      ▷ Call HCD
    $R \leftarrow$ HIR($A$)                                                      ▷ Call HIR
    **return**
  **end procedure**
  **procedure** HCD($D$)
    $D' \leftarrow$ PREPROCESSING($D$) ▷ Remove links, hyperlinks, '@', newline, punctuations (except '.', ',' and '?') and labels to unique numbers
    $e \leftarrow$ BERT_Encoder($x$)
    $p \leftarrow$ softmax($Linear(e[CLS])$)
    $c \leftarrow$ argmax($p$)
    $R \leftarrow$ None
    **if** If c is 1 **then**
        Let b is the baseline text for x
        $x\_embed \leftarrow BERT(embedding\_layer)(x)$
        $b\_embed \leftarrow BERT(embedding\_layer)(b)$
        Let the number of steps is n.
        Calculate value of alpha using $alpha \leftarrow [0, 1, \frac{1}{n}]$.
        Initialize $sum\_of\_gradients \leftarrow 0$
        **for** $\alpha \in$ alpha **do**
            Calculate the interpolated embedding $i\_e$, i.e. $i\_e \leftarrow x\_embed + \alpha * (x\_embed - b\_embed)$
            Calculate $\nabla i\_e$ w.r.t $p$
            $sum\_of\_gradients \leftarrow sum\_of\_gradients + \nabla i\_e$
        **end for**
        $IG(x) \leftarrow \frac{sum\_of\_gradients}{n}$
        Normalize the $IG(x)$ along the embedding dimension to get the attribution score for each token.
        Sum the attribution score of subword tokens to get the attribution for words.
        Calculate the top-k (here, k=5) tokens in descending order of attribution score.
        R ← top-k
    **else**
        x is detected non-hate
    **end if**
  **end procedure**
  **procedure** HIR($x, A, R$)
    **for** word $\in$ x **do**
        **if** word in A **then**
            Replace word ← ['MASK']
        **end if**
    **end for**
    Pass this x' to the BERT model and predict the word for these masked words.
    For each masked word, select a list of the top 20 words (maybe 10).
    In this way, we have 20 sentences to replace the hate text.
    Now, calculate the similarity score for each sentence and choose the one having the highest similarity value.
    **return**
  **end procedure**

---

## 5.2 HASOC dataset

$HASOC(2019)_{Hindi}$ [36], $HASOC(2020)_{Hindi}$ [38] and $HASOC(2021)_{Hindi}$ [39] is sampled from ™Twitter and ™Facebook using hashtags and keywords. In these datasets, we have to classify the tweets into two classes (Sub-task A is a coarse-grained binary classification): hate and offensive (HOF) and non-hate (NOT). HOF indicates that a post contains hate speech, offensive language, or both. NOT implies no hate speech or other offensive material in this post. Training and test data are provided separately. $HASOC(2021)_{Marathi}$ [39]

### 5.3 MACD dataset

$MACD_{Hindi}$, $MACD_{Tamil}$ and $MACD_{Telegu}$ dataset [62] is released by ShareChat in collaboration with CNERG Lab, IIT Kharagpur, which is well-balanced and human-annotated, where comments have been sourced from a popular social media platform - ShareChat. MACD contains the training, validation, and test split in CSV format for all the languages included with MACD - Hindi, Tamil, Telugu, Malayalam, and Kannada. The dataset contains two labels - 0 (for abusive comments) and 1 (for non-abusive comments). We Used the Hindi, Tamil and Telegu dataset.

### 5.4 BD-SHS dataset

$BD-SHS_{Bengali}$ dataset [61] is created by collecting the Bengali article from various sources, including a Bengali Wikipedia dump, Bengali news articles like Daily Prothom Alo, Anandbazar Patrika, BBC, news dumps of TV channels (ETV Bangla, ZEE news), social media (™Twitter, ™Facebook pages and groups, ™LinkedIn), books, and blogs. The raw text corpus consists of 250 million articles. This dataset consists of 30,000 instances, where 10,000 instances belong to the hate category, and 20,000 instances belong to non-hate. Hates are further classified as political, personal, gender abusive, geopolitical, or religious hate.

| Dataset | Model | Accuracy | Precision | Recall | Macro F1 |
|---|---|---|---|---|---|
| $HASOC(2019)_{Hindi}$ | **google/muril-base-cased** | **0.84 ± 0.010** | **0.84 ± 0.010** | **0.84 ± 0.010** | **0.84 ± 0.010** |
| | l3cube-pune/hindi-bert-v2 | 0.84 ± 0.001 | 0.84 ± 0.001 | 0.84 ± 0.001 | 0.84 ± 0.001 |
| $HASOC(2021)_{Marathi}$ | google/muril-base-cased | 0.88 ± 0.012 | 0.87 ± 0.012 | 0.87 ± 0.010 | 0.87 ± 0.012 |
| | **l3cube-pune/marathi-roberta** | **0.90 ± 0.011** | **0.89 ± 0.011** | **0.88 ± 0.011** | **0.88 ± 0.011** |
| $MACD_{Hindi}$ | **google/muril-base-cased** | **0.86 ± 0.002** | **0.86 ± 0.003** | **0.86 ± 0.003** | **0.86 ± 0.003** |
| | l3cube-pune/hindi-bert-v2 | 0.86 ± 0.002 | 0.86 ± 0.003 | 0.86 ± 0.003 | 0.86 ± 0.003 |
| $MACD_{Tamil}$ | google/muril-base-cased | 0.88 ± 0.004 | 0.88 ± 0.005 | 0.88 ± 0.005 | 0.88 ± 0.005 |
| | **l3cube-pune/tamil-bert** | **0.88 ± 0.001** | **0.88 ± 0.001** | **0.88 ± 0.001** | **0.88 ± 0.001** |
| $MACD_{Telegu}$ | google/muril-base-cased | 0.89 ± 0.003 | 0.89 ± 0.005 | 0.89 ± 0.005 | 0.89 ± 0.005 |
| | **l3cube-pune/telugu-bert** | **0.90 ± 0.001** | **0.89 ± 0.001** | **0.90 ± 0.001** | **0.90 ± 0.001** |
| $BD-SHS_{Bengali}$ | google/muril-base-cased | 0.92 ± 0.001 | 0.92 ± 0.001 | 0.92 ± 0.001 | 0.92 ± 0.001 |
| | **l3cube-pune/bengali-bert** | **0.92 ± 0.001** | **0.92 ± 0.001** | **0.92 ± 0.001** | **0.92 ± 0.001** |

**Table 2** Precision, Recall, and F1 score on all five datasets for hate speech detection task

### 5.5 settings

We have reported all results after 5-fold cross-validation. At the training time, early stopping is used as one of the stopping criteria. All hyper-parameters are tuned on the validation partition of each dataset.

For training optimization, we mostly use the following details—

1. Optimizer: AdamW
2. Momentum = 0.9, with Learning rate = 1e-5
3. Learning rate scheduler: linear-scheduler with warm-up steps
4. Loss function: Cross-Entropy loss

All pre-trained models are incorporated from https://huggingface.co/ [63].

## 5.6 Results

In this section, we scribe all experiment results and analyses based on the three tasks—hate speech detection, hate speech identification, and hate speech reduction. For the hate speech detection task, we compare several state-of-the-art transformer-based text classification models for automatic metric-based evaluations. Regarding hate speech identifications, we provide both results generated by IG and human evaluation. For the hate speech reduction task, we evaluate model performance solely based on human evaluation. Human annotations are often considered gold-standard for almost every NLP task. In numerous NLP applications, from machine translation to fact-checking, several past studies [64] have already depicted that automatic metric-based evaluations are insufficient. Although human-based evaluation requires much effort and significant evaluation time compared to automatic metric-based evaluations, a well-designed human evaluation over a small sample of model outputs attests to model accuracy for real-world applications. Due to these facts, we incorporate a human-based evaluation in our experiments.

### 5.6.1 Hate Context Detection

For hate speech detection task evaluation, four automatic metrics are used in our experiments: accuracy, precision, recall, and F1 score. A micro-average version of these four metrics is considered as we observed class imbalance in our incorporated datasets. Table 2 shows the performance of three different transformer-based hate speech detection models: Google-MuRIL, Indic-BERT, and XLM-RoBERTa. For each model's output over every dataset, we compute all four metrics, along with dispersion (standard deviation) across cross-validation folds. Table 2 suggests the Google-MuRIL model as the top-scoring model for hate-detection tasks over all five datasets. Although Indic-BERT and XLM-RoBERTa perform almost similarly, we find that XLM-RoBERTa yields slightly better results (table 2). It is important to note that Google-MuRIL not only depicts the best results but also establishes lower dispersion than the other two models over almost all datasets.

For evaluating the hate speech identification task, we pick positive attributed hate words from a given context (sentence) and match those picked words based on the ground truth. The similarity between predicted hate words and ground truth hate words is calculated through the Jaccard Index [65]. In human-based evaluation for hate speech identification, we also compute the Jaccard Index between model-predicted hate words and human-identified hate words.

**Human-based Evaluation.** For hate speech identification through human evaluation, we selected three graduate students from different departments (with dissimilar backgrounds) to ensure diversity among the human annotator's pool. All annotations from these annotators are collected in a normal lab environment based on given instructions. Table 3 shows the IG as well as human evaluation.

We considered a sample of 40 sentences from each instance where our proposed method detects at most three hate words. Each annotator is asked to determine three topmost hate words from each sample sentence. Finally, we obtain 0.5 Jaccard index similarity between model-predicted and human-detected hate words. The Jaccard index value indicates that the proposed method performs similarly to humans on the hate speech identification test.

| Text | Words with positive attribution scores by Integrated Gradients | Human Evaluation |
|---|---|---|
| तेरी *ड में प्याज काट देगा गुज्जर **ड़ी के | *ड, **ड़ी, में | *ड, काट, **ड़ी |
| इस रशीदी जेहादी का पायजामा पीला तो नही है आज आज ह** को बड़ा भाईचारा नज़र आ रहा है | जेहादी, रशीदी, भाईचारा | रशीदी, जेहादी, ह*** |
| यह लड़कियों के नाम पर कलंक है बहन की लो** उनके घर का स्टेटस भी ऐसा ही होगा इनके जैसे उनकी मम्मी भी कहीं जाती होगी | बहन, लो***, लड़कियों | कलंक, बहन, की, लो** |

**Table 3** Words with positive attribution scores by Integrated Gradients and human annotation

### 5.6.2 Hate Intensity Reduction

For the hate intensity reduction, positive attributed hate words are replaced with the special token—"[MASK]". We generated a set of candidates by replacing those masked hate words with different high probable words based on contexts, using the MLM concept. To select an optimal candidate for the candidate set, we use the BERTScore as a sentence similarity metric. As replacing hate words in a sentence with another highly probable word (yield modified sentence) doesn't always reduce the degree of hate content of the sentence, the incorporation of a proper faithful assessment is crucial for our proposed method. We perform a human-based evaluation for this hate intensity reduction task with the same setup which we use for the hate speech identification task, We gave a sample of 40 modified sentences to those human annotators and asked them to rate them on a 1-5 Likert scale, where 5 denotes the absolute reduction of hate content from hate sentences. We obtain average scores of 4.1 (out of 5), 4.2 (out of 5) and 3.8 (out of 5) corresponding to the three annotators w.r.t. the Likert scale. Statistically, all annotators have a $r = 0.86$ degree of agreement (Pearson correlation). High levels of annotator agreement and high average scores per annotator indicate that our proposed model for reducing hate speech performs well.

## 6 Error Analyses

Although our proposed method shows superior performances in both hate word identification and hate content reduction, in some instances, our proposed method fails to produce legitimate outputs. We perform error analyses for two subtasks—i. hate words identification. ii. hate reduction task.

## 6.1 Hate Words Identification

Even though the proposed model shows promising results in hate word identification, we discover a limitation. The proposed model often shortfalls in identifying all hate words within a given sentence where many hate words are present. In the following case, we have shown that even after detecting hate words from the given sentence, several hate words still exist.

- Given input: "बस एक ही शब्द है क*ला *ड की पैदाइश है मा***द इसकी माँ को मौलाना ने *क *क के इतना गन्दी औलाद पैदा किया है मा⎵!*द".
- Hate words (red colored): "बस एक ही शब्द है क*ला *ड की पैदाइश है मा***द इसकी माँ को मौलाना ने *क *क के इतना गन्दी औलाद पैदा किया है मा***द".
- Detected hate words (green colored): "बस एक ही शब्द है क*ला *ड की पैदाइश है मा***द इसकी माँ को मौलाना ने *क *क के इतना गन्दी औलाद पैदा किया है मा***द".

## 6.2 Hate Intensity Reduction

In the hate reduction subtask, we find that the proposed model poses two types of errors.

- **Position of hate words in sentence:** Our proposed method's performance in reducing hate intensity significantly varies with hate word positions in a sentence. If a detected hate word appears at the beginning of a sentence, our proposed method often fails to reduce hate intensity. For example, the sentence "सा* रं* गोरखपुर को बदनाम करती" is reduced to "दलाल टीम गोरखपुर को बदनाम करती". Red parts denote the hate content. So, we can observe that even after hate intensity reduction, the resultant sentence is still showing hate content.
- **Coagulation of multiple hate words in sentence:** Coagulation of multiple hate words in a sentence also deteriorates the proposed method's credibility while reducing hate intensity content. Several observations on our proposed method outcomes show that if multiple hate words appear together in sentences, then our proposed method fails to reduce hate intensity from those sentences. For example, the sentence "आपकी खु**ली मिटाने के लिये हम *ल ल* *र तयार है" is reduced to "आपकी दर्द मिटाने के लिये हम *ड मारने को तयार है". This demonstrates that the proposed method is ineffective in reducing hate speech when multiple hate words appear together.

# 7 Conclusion and Future Work

In this paper, we intend to build an end-to-end deep learning-based system that detects hate speech, identifies the top-k words, and reduces the hate intensity from the hate text. We introduce a novel approach for reducing hate intensity from a given sentence. According to our knowledge, the presented work is the first attempt at hate speech reduction in the field of hate/offensive/abusive text analyses. This work doesn't need any domain knowledge-based pre-processing and expert tuning. The extensive result section combined with both automatic metric-based evaluations and human-based evaluations indicates the stellar performance of our proposed method. Towards the end of the paper, we also present some caveats of our proposed method in the error analysis section. In the future, we will concentrate on solving those caveats depicted by our proposed method. Reducing the hate intensity of material in broad discourses should be a significant future direction for our work.

## 8 Acknowledgement

This research is funded by the Science and Engineering Research Board (SERB), Dept. of Science and Technology (DST), Govt. of India through Grant File No. SPR/2020/000495.